\documentclass[letterpaper]{article} % DO NOT CHANGE THIS
\usepackage[]{aaai23}  % DO NOT CHANGE THIS
\usepackage{times}  % DO NOT CHANGE THIS
\usepackage{helvet}  % DO NOT CHANGE THIS
\usepackage{courier}  % DO NOT CHANGE THIS
\usepackage[hyphens]{url}  % DO NOT CHANGE THIS
\usepackage{graphicx} % DO NOT CHANGE THIS
\usepackage{natbib}  % DO NOT CHANGE THIS AND DO NOT ADD ANY OPTIONS TO IT
\usepackage{caption} % DO NOT CHANGE THIS AND DO NOT ADD ANY OPTIONS TO IT
\usepackage{algorithmic}
\usepackage{color}
\usepackage{colortbl}
\usepackage{pifont}

\usepackage{pifont}
\usepackage{epsfig}
\usepackage{amsmath}
\usepackage{amssymb}
\usepackage{booktabs}
\usepackage{xcolor}
\usepackage{enumitem}
\usepackage{lipsum}
\usepackage{multirow}
\usepackage{bbding}
\usepackage{subfigure}

\usepackage[misc]{ifsym}

\usepackage[ruled,vlined,linesnumbered]{algorithm2e}
\makeatletter
\newcommand{\algorithmfootnote}[2][\footnotesize]{%
  \let\old@algocf@finish\@algocf@finish% Store algorithm finish macro
  \def\@algocf@finish{\old@algocf@finish% Update finish macro to insert "footnote"
    \leavevmode\rlap{\begin{minipage}{\linewidth}
    #1#2
    \end{minipage}}%
  }%
}
\makeatother

\newcommand{\app}{\raise.17ex\hbox{$\scriptstyle\sim$}}

\usepackage{newfloat}
\usepackage{listings}
\DeclareCaptionStyle{ruled}{labelfont=normalfont,labelsep=colon,strut=off} % DO NOT CHANGE THIS
\nocopyright % -- Your paper will not be published if you use this command
\title{Quality Matters: Embracing Quality Clues for Robust 3D Multi-Object Tracking}
\author{
    Jinrong Yang\textsuperscript{\rm 1}\equalcontrib,
    En Yu\textsuperscript{\rm 1}\equalcontrib,
    Zeming Li\textsuperscript{\rm 2},
    Xiaoping Li\textsuperscript{\rm 1}\footnotemark[2],
    Wenbing Tao\textsuperscript{\rm 1}\thanks{Corresponding authors}
}
\affiliations{
    \textsuperscript{\rm 1}Huazhong University of Science and Technology, \textsuperscript{\rm 2}MEGVII Technology\\
    \{yangjinrong, yuen, lixiaoping, wenbingtao\}@hust.edu.cn, lizeming@megvii.com
}

\usepackage{bibentry}
\begin{document}

\maketitle

\begin{abstract}
3D Multi-Object Tracking (MOT) has achieved tremendous achievement thanks to the rapid development of 3D object detection and 2D MOT. Recent advanced works generally employ a series of object attributes, \emph{e.g.,} position, size, velocity, and appearance, to provide the clues for the association in 3D MOT. However, these cues may not be reliable due to some visual noise, such as occlusion and blur, leading to tracking performance bottleneck. To reveal the dilemma, we conduct extensive empirical analysis to expose the key bottleneck of each clue and how they correlate with each other. The analysis results motivate us to efficiently absorb the merits among all cues, and adaptively produce an optimal tacking manner. Specifically, we present \textit{Location and Velocity Quality Learning}, which efficiently guides the network to estimate the quality of predicted object attributes. Based on these quality estimations, we propose a quality-aware object association (QOA) strategy to leverage the quality score as an important reference factor for achieving robust association. Despite its simplicity, extensive experiments indicate that the proposed strategy significantly boosts tracking performance by 2.2\% AMOTA and our method outperforms all existing state-of-the-art works on nuScenes by a large margin. Moreover, QTrack achieves 48.0\% and 51.1\% AMOTA tracking performance on the nuScenes validation and test sets, which  significantly reduces the performance gap between pure camera and LiDAR based trackers.
\end{abstract}

\section{Introduction}
\label{introduction}

3D Multi-Object Tracking (MOT) has been recently drawing increasing attention since it is widely applied to 3D perception scenes, e.g., autonomous driving, and automatic robot. The 3D MOT task aims at locating objects and associating the targets of the same identities to form tracklets. According to the used sensors, existing 3D MOT methods can mainly be categories into two classes, i.e., camera-based and LiDAR-based schemes. In this paper, we mainly delve into the camera-only scheme since it contains semnatic information and is more economical.

\begin{figure}[t]
\centering
\includegraphics[width=1.0\columnwidth]{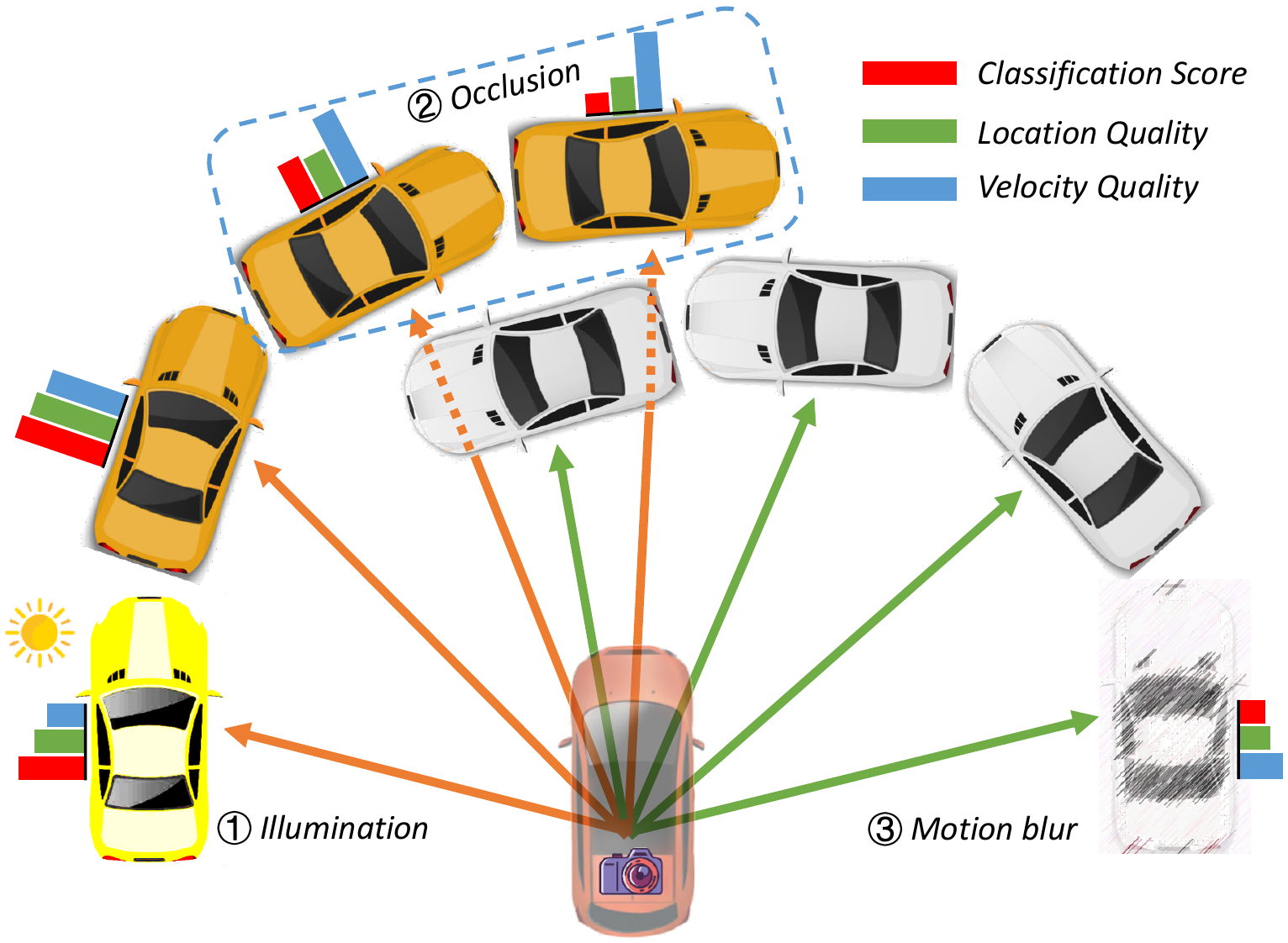}
\caption{Illustration of three type of hard cases: (1) illumination of external, (2) occlusion, (3) motion blur. The red, green and blue pillars are organized to represent the classification score, location quality, and velocity quality, where the higher pillars indicate higher values.}
\label{fig:motivation}
\end{figure}

Existing 3D MOT methods mostly adopt the tracking-by-detection paradigm. In this regime, a 3d detector is firstly employed to predict 3D boxes and the corresponding classfication scores, and then some post-processing methods (e.g., motion-based~\cite{kalman} or appearance-based) are used to line detected targets to form trajectories. In the camera scheme, it is natural to extract objects' discriminative appearance features~\cite{deft,quasi-dense} to represent targets and use the features to measure the similarities among detected targets. However, the procedure of extracting the appearance feature is cumbersome since it requires predicting high-dimensional embedding, which is hard for joint training due to the optimization contradiction between the detection and embedding branches \cite{yu2022relationtrack}. Moreover, it is difficult to deal with the notorious occlusion and motion blur issues. Some other methods~\cite{ab3d, simpletrack} build a motion model (Kalman Filter) to obtain some desired states of tracking clues (e.g, center position, size, size ratio, or rotation) by a linear motion assumption. Nevertheless, this process involves various hyper-parameters (e.g., initialization uncertainty of measurement, state and process, etc.) and executes complex matrix transpose operation. Different from the aforementioned methods, CenterPoint~\cite{centerpoint} reasonably leverages predicted center locations and velocities of targets for building motion. In detail, it uses time lag between two moments of observations to multiply the predicted velocity for linear location prediction. Afterwards, the L2 distance among targets acts as a measurement metric for the association procedure. For simplicity, we call this tracking framework CV method. It shows effectiveness to achieve remarkable tracking performance, while only conducting a simple operation (i.e., matrix addition and multiplication) for parallel cost computation.

\begin{figure*}[t]
\subfigure[Location Quality Distribution]{
    \includegraphics[width=0.32\linewidth]{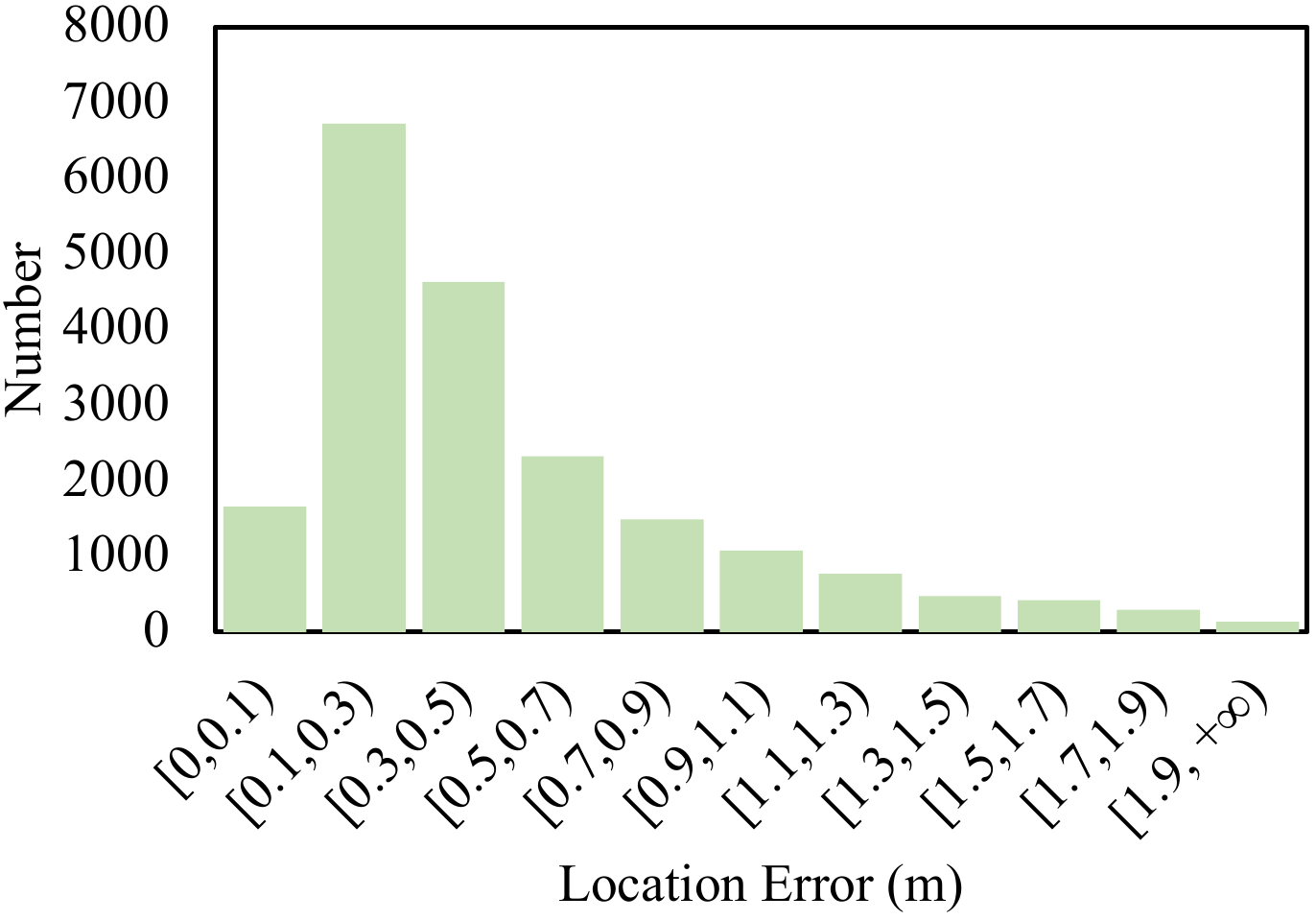}
    \label{fig:analyse_1}
    }
    \subfigure[Velocity Quality Distribution]{
    \includegraphics[width=0.318\linewidth]{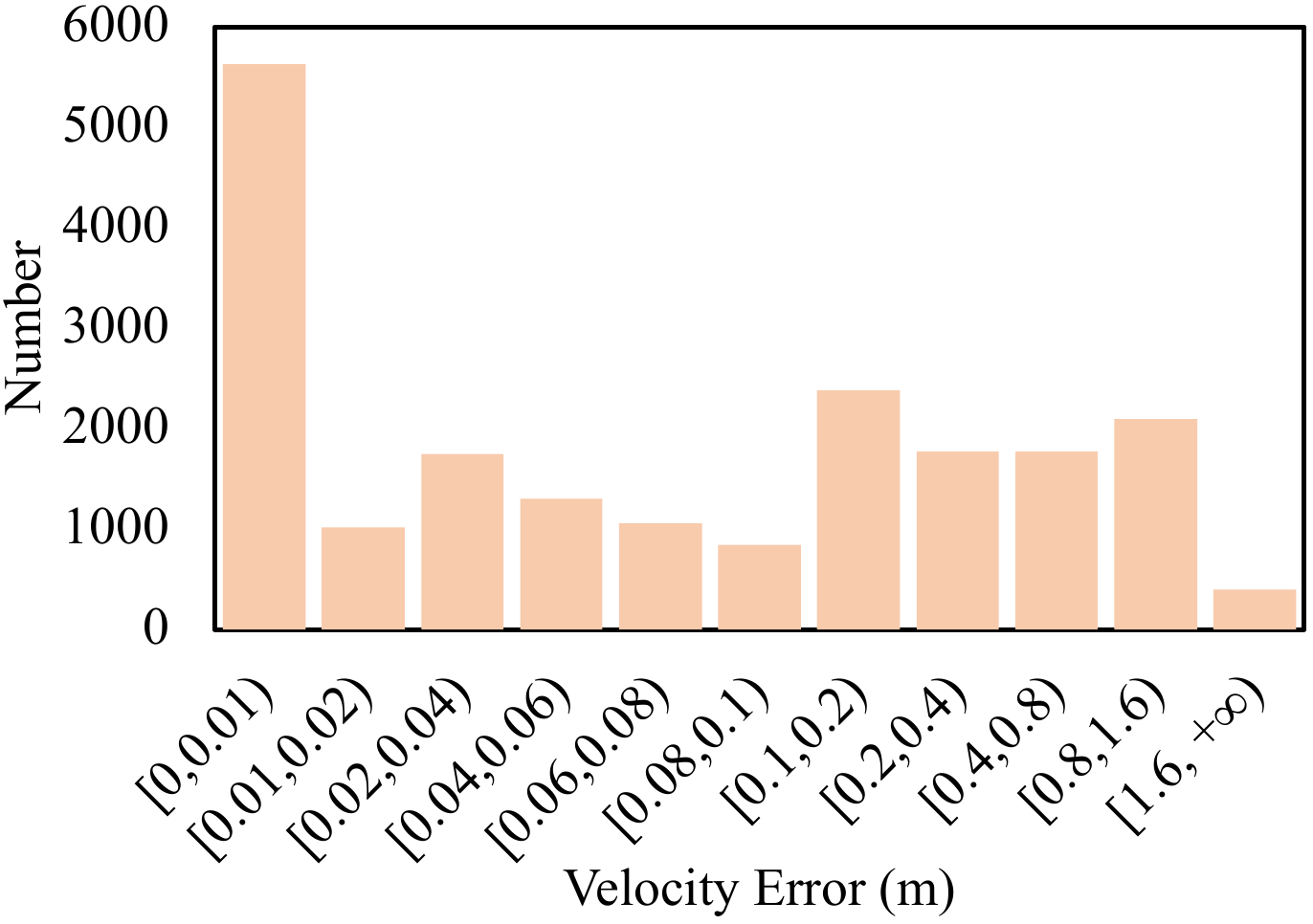}
    \label{fig:analyse_2}
    }
    \subfigure[Correlation]{
    \includegraphics[width=0.31\linewidth]{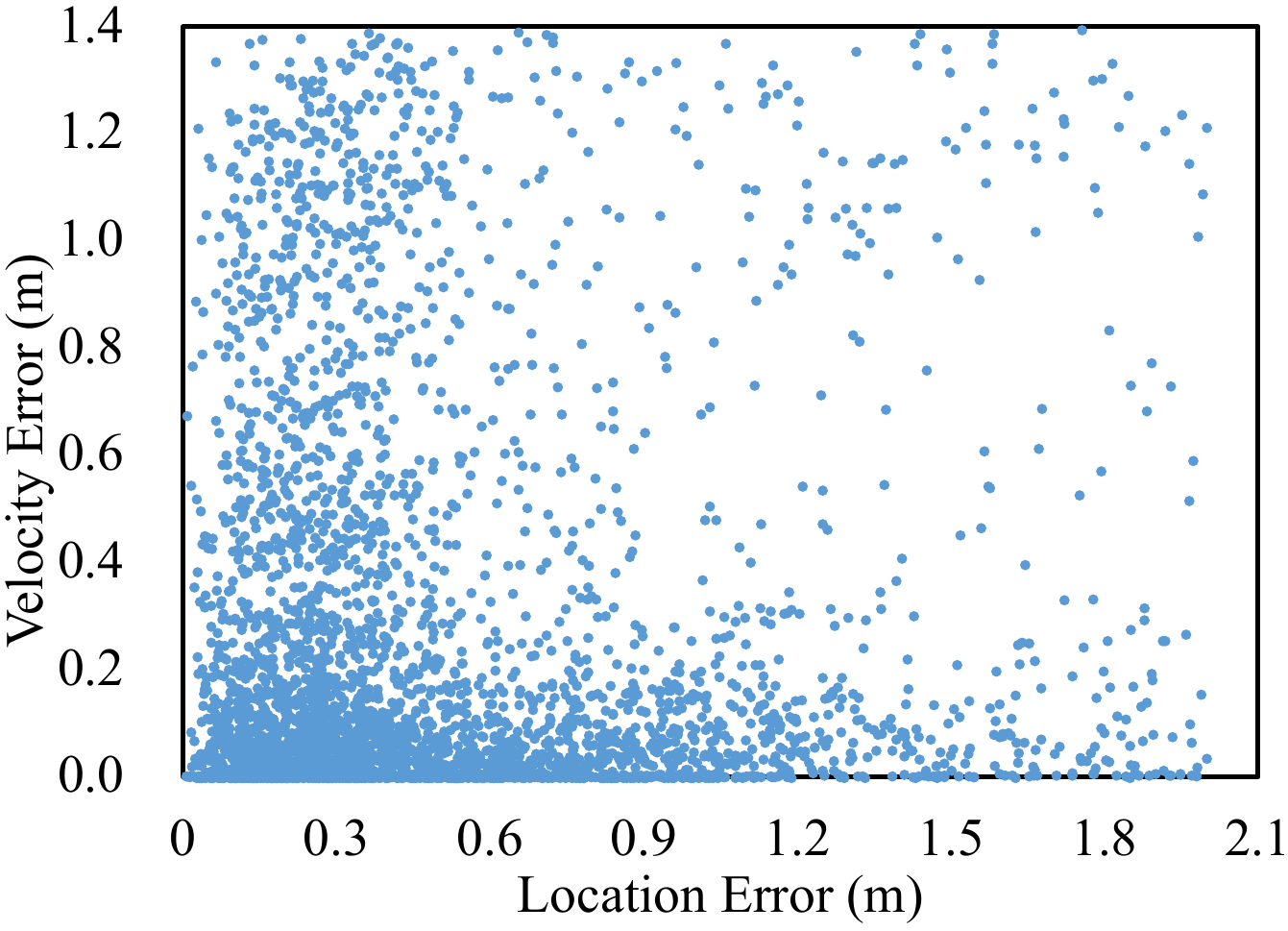}
    \label{fig:analyse_3}
    }
\centering
% \vspace{-1em}
\caption{Statistics of location and velocity quality distribution and their correlation on nuScenes \emph{val} dataset. (a) and (b) reveal that the accuracy of all prediction location and velocity are varying on an unfixed scale, showing irregularity. (c) shows a messy scatter plot, which reflects no relations between location and velocity results.}
\label{fig:analyse}
\end{figure*}

Although the CV framework shows efficiency for 3D MOT task, it relies heavily on the predicted quality of center location and velocity. The requirement may be harsh for the 3D base detector, since estimating the center location and velocity of an object from a single image is exactly an ill-posed problem. As shown in Fig. 1, notorious occlusion, motion blur, and the illumination of external issues will significantly disturb the estimation performance. To further confirm this issue, we conduct an empirical analysis to study the predicted center location and velocity quality distribution as well as their correlations. Our study reveals two valuable points: (1) There exists a significant gap between the estimation error of 3D centers and that of velocities; (2) The predicted quality of location and velocity is extremely misaligned. The imbalanced tracking cues have little effect on the detection performance but play a dramastical role in MOT. The analysis cues motivate us to endow each predicted box with the self-diagnosis ability to tracking clues for realizing stable tracking association.

To this end, we propose to forecast the quality of tracking clues from the base 3D detector. Specifically, we introduce a Normalized Gaussian Quality (NGQ) metric with two dimensions to measure the quality of predicting center location and velocity. NGQ metric comprehensively considers the vector errors of the two predictions in a 2D vector space, which is a prerequisite for our tracking framework. Based on the quality estimation of NGQ, we design a robust association mechanism, i.e, Quality-aware Object Association (QOA) strategy. It adopts the velocity quality to filter out low quality motion candidates, and leverages the location quality to further rule out center positions of boxes with bad estimations. Therefore, QOA not only effectively deals with hard cases, but also avoids dangerous association. In a sense, our method is subordinate to the idea of "Put Quality Before Quantity" principle.

Through combining the proposed methods with the baseline 3D detector, we obtain a simple and robust 3D MOT framework, namely \textit{quality-aware 3D tracker} (QTrack). We conduct extensive experiments on nuScenes dataset~\cite{nuscenes}, showing significant improvements in the 3D MOT task. Comprehensively,  the contributions of
this work are summarized as follows:
\begin{itemize}
\item We conduct extensive empirical analysis to point out that the predicted quality of center location and velocity exist large distribution gap and misalignment relationship, making efficient CV tracking framework fall into sub-optimal performance.

\item We first propose to predict the quality of velocity and location quality measured by our designed NGQ metric. Afterwards, we further introduce QOA to leverage the two qualities for insuring safe association in 3D MOT task.

\item The overall 3D MOT framework (QTrack) achieves SOTA performance on nuScenes dataset which outperforms other camera-based methods by a large margin. Specially, QOA improves the baseline tracker by +2.2\% AMOTA among several 3D detector settings, showing its effectiveness. 
\end{itemize}

\section{Related Work}
\label{relative_work}
\subsection{3D Multi-object Tracking}
Thanks to the development of 3D detection \cite{huang2021bevdet, li2022bevdepth, liu2022petr} and 2D MOT technologies \cite{han2022mat, yu2022relationtrack, yu2022towards, bytetrack}, recent 3D MOT methods \cite{ab3d, centerpoint, deft, quasi-dense, simpletrack} mainly follow tracking-by-detection paradigm. These trackers following this paradigm first utilize a 3D object detector to localize the targets in the 3D space (including location, rotation, and velocity) and then associate the detected objects with the trajectories according to various cues (location or appearance).

Traditional 3D MOT usually uses a motion model (Kalman filter) to predict the location of the tracklets and then associate the candidate detections using 3D (G)IoU \cite{ab3d,simpletrack} or L2 distance \cite{centerpoint}. Some works also utilize advanced appearance model (ReID) \cite{deft, gnn3dmot, deft} or temporal model (LSTM) \cite{triplettrack,quasi-dense} to provide more reference cues for the association. Recently, Transformer \cite{transformer} has been used in 3D detection \cite{detr3d} and MOT \cite{time3d, mutr3d} to learn 3D deep representations with 2D visual information and trajectory encoded. Although these methods achieved remarkable performance, when they are applied to complex scenarios (e.g., occlusion, motion blur, or light weakness), the tracking performance becomes unsatisfactory. In this work, we argue that a simple velocity clue with quality estimation can deal with the corner cases and achieve robust tracking performance. Our proposed QTrack focuses on how to assess the quality of the location and velocity prediction, and then make full use of these quality scores in the matching process.

\subsection{Prediction Quality Estimation}
To estimate the quality of model's prediction is a non-trivial, which can be applied to tackle prediction imbalance or decision-making. In the field of object detection, advanced works~\cite{fcos3d,fcos,iounet} introduce to predict a box's centerness or IoU for perceiving the quality of prediction (3D) boxes. \cite{maskscoring} employ the method to perceive the mask predicted quality. These methods can alleviate the imbalance between classification score and location accuracy. \cite{li2022diversity} introduces a uncertainty based method to estimate the predicted quality of several depth factors, and then the quality is employed to make optimal decisions. In this paper, we introduce to predict the predicted quality of velocity and location. Afterwards, the predicted quality will be adopted to eliminate the non-robust association case of tracking task. To our knowledge, our work is the first effort to perceive the quality velocity and location for the decision-making in 3D MOT task.

\subsection{Multi-View 3D Object Detection}
3D object detection is the predecessor task for 3D MOT task. It can be split into two stream methods including point-based detectors~\cite{pointpillars, second, centerpoint, pointrcnn, pvrcnn, dbq} and camera-based ones~\cite{fcos3d, huang2021bevdet, li2022bevdepth, detr3d, liu2022petr, bevformer}. In this paper, we focus on the 3D MOT for the multi-view camera based framework, which has made tremendous advance. Transformer based methods~\cite{detr3d, liu2022petr, bevformer} introduce 3D object queries to interact with the multi-view image feature map. 3D object queries are constantly refined to predict 3D boxes and other tasks in an end-to-end manner. BEVDet~\cite{huang2021bevdet} and BEVDepth~\cite{li2022bevdepth} directly project the multi-view image feature into BEV representation and attach a center-based head~\cite{centerpoint} to conduct detection task. Standing on the shoulders of giants, we aim to equip BEVDepth with the ability to perceive the quality of velocity and center locatopn, which is the key to diagnose non-robust association for tracking. Then we introduce a novel “tracking by detection” (QTrack) to endow BEVDepth with effiective and efficient tracking ability.

\section{Methodology}

\subsection{Delve into the Quality Distribution}

We aim to solve the task of 3D multi-object tracking (3D MOT), the goal of which is to locate the objects in the 3D space and then associate the detected targets with the same identity into the tracklets. The key challenge is how to associate the tracklets efficiently and correctly. In contrast to the motion-based and appearance-based association strategies, we argue that the simple velocity clue (CV method) is enough for the association, which is more lightweight and deployment-friendly. However, the performance of the existing CV tracking framework is not satisfactory. To analyze the reason for the limited performance of tracking with velocity, we count and visualize the distribution of the prediction error between location and velocity. As illustrated in Fig. 2 (a) and (b), we can observe that the distribution of the location and velocity quality (prediction error) is scattered, and a sizable number of low-quality boxes are included. Moreover, Fig. 2 (c) shows that the distribution correlation between the location and velocity error is nonlinear, which means the quality of the location and velocity is seriously misaligned.

Based on these observations, we conclude that the limited performance of tracking with velocity is due to the following reasons: (1) Low quality of the location or velocity. When one of the location and velocity predictions is not accurate enough, the tracker can not perform well even if the other prediction is reliable. (2) Misalignment between the quality of location and velocity. We should take both location and velocity quality into consideration. Driven by this analysis, we propose \textit{Location and Velocity Quality Learning} to learn the quality uncertainty of the location, and velocity that can assist the tracker to select high-quality candidates for the association.

\subsection{Base 3D Object Detector}

Our method can be easily coupled with most existing 3D object detectors with end-to-end training. In this paper, we take BEVDepth~\cite{li2022bevdepth} as an example. BEVDepth is a camera-based Bird's-Eye-View (BEV) 3D object detector that transfers the multi-view image features to the BEV feature through a depth estimation network and then localizes and classifies the objects in the BEV view. It consists four kinds of modules: an image-view encoder, a view transformer with explicit depth supervision utilizing encoded intrinsic and extrinsic parameters, a BEV encoder and a task-specific head. The entire network is optimized with a multi-task loss function:

\begin{equation}
\mathcal{L}_{det} = \mathcal{L}_{depth} + \mathcal{L}_{cls} + \mathcal{L}_{reg},
  \label{Eq1}
\end{equation}
where the depth loss $\mathcal{L}_{depth}$, classification loss $\mathcal{L}_{cls}$ and regression loss $\mathcal{L}_{reg}$ remain the same setting as the original paper. As illustrated in Fig. 3, the task of the regression branch includes heatmap, offsets, height, size, rotation and velocity.

\begin{figure}[t]
\centering
\includegraphics[width=1.0\columnwidth]{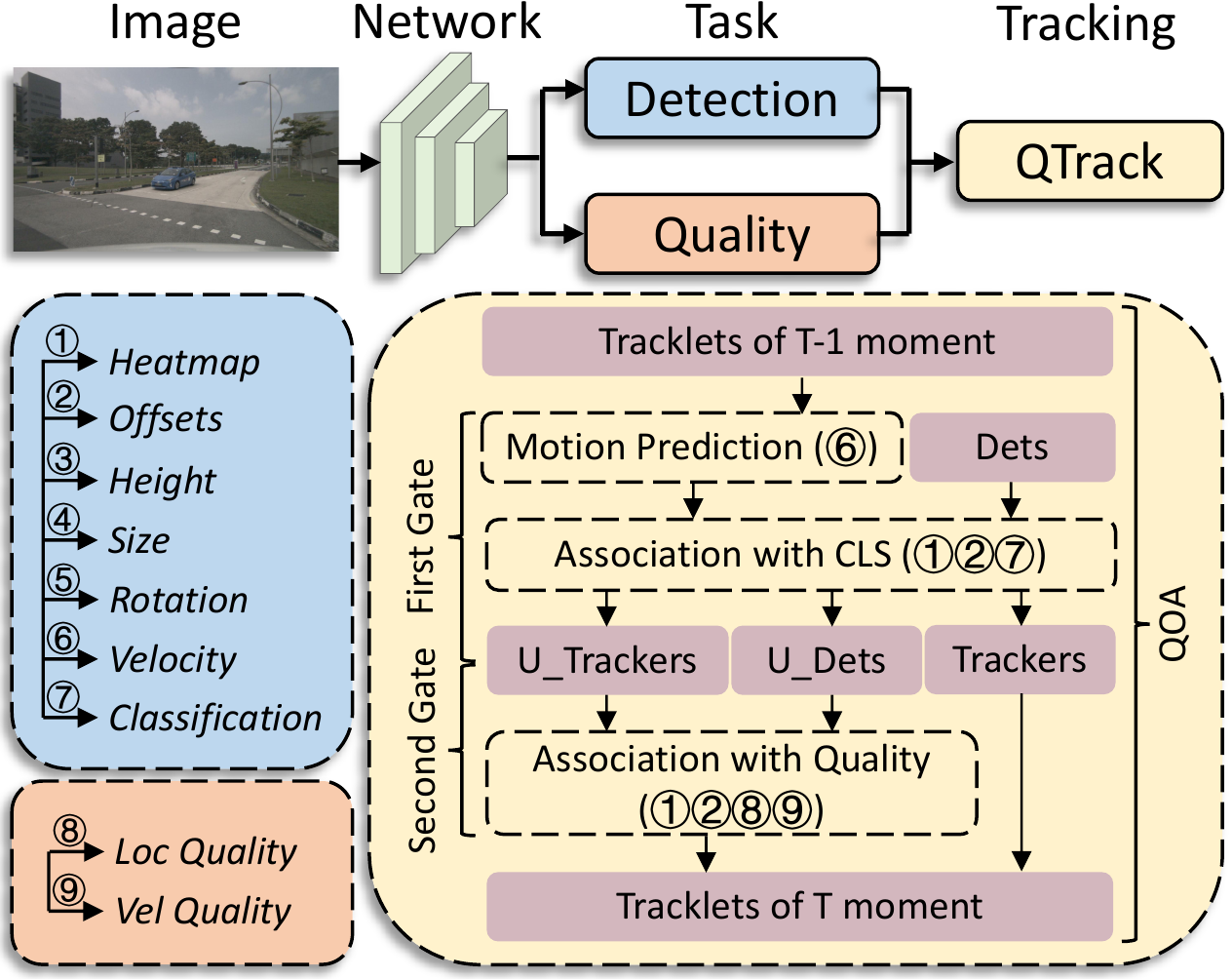}
\caption{Overview of base 3D detector and QTrack. The multi-view images are first fed into detector, i.e., BEVDepth. Then we add two parallel branches for predicting location and velocity quality, respectively. For QTrack, it first employs velocity clue to conduct motion predicted, and then adopts heatmap, offsets, and classification score to carry out association procedure in the first gate stage. Specially, location and velocity qualities are introduced to execute this work's key module QOA for unmatched trackers and detections in the second gate stage.}
\label{fig:pipeline}
\end{figure}

\subsection{Location and Velocity Quality Learning}

To effectively estimate the quality of location and velocity, it first needs to define the quality measurement metric. Technically, the box's center location is calculated by incorporating predicted heatmap and corresponding offsets so that the location quality can be simplified to offset predicted quality. Specially, the offsets and velocity are defined in a 2-dimensional vector space. We introduce a Normalized Gaussian Quality (NGQ) metric to represent their quality. Given a predicted vector $\mathbf{P} \in \mathbb{R}^{2}$ and ground truth vector $\mathbf{G} \in \mathbb{R}^{2}$, we formulate NGQ metric as:

\begin{equation}
\begin{aligned}
\mathrm{NGQ} = e^{-{\frac{\sqrt{{(\mathrm{P}_x - \mathrm{G}_x)^2 + (\mathrm{P}_y - \mathrm{G}_y)^2}}}{\gamma}}},
\label{Eq2}
\end{aligned}
\end{equation}
where the subscripts $x$ and $y$ indicate the value in the x and y directions while $\gamma$ is a hyper-parameter to control the value distribution of NGQ. We set $\gamma$ to 1.0 and 3.0 for location and velocity, respectively. $\mathbf{P}$ and $\mathbf{G}$ can be instantiated as predicting offset and velocity. When the prediction is equal to ground truth, NGQ = 1, while the predicted error is larger, NGQ is closer to 0.

After defining the quality, we elaborate on how to learn it. As shown in Fig. 3, we attach a $3 \times 3$ convolution layer for offset and velocity branch to predict location quality $\mathrm{NGQ}^{loc} \in \mathbb{R}^{1}$ and velocity quality $\mathrm{NGQ}^{vel} \in \mathbb{R}^{1}$, respectively. The quality supervision is conducted by binary cross entropy (BCE) loss:

\begin{equation}
\begin{aligned}
& \mathcal{L}_{quality} = -\frac{1}{N}\sum_{i=1}^{N}[{{\hat{\mathrm{NGQ_i}} \cdot \log{\mathrm{NGQ_i}}}} \\ 
& + {(1 - \mathrm{NGQ_i}) \cdot \log{(1 - \hat{\mathrm{NGQ_i}})}}],
\label{Eq3}
\end{aligned}
\end{equation}
where $\mathrm{\hat{NGQ}}$ is the ground truth quality calculated by Eq.~\ref{Eq2}. This far, the total loss for our detector is formulated as:

\begin{equation}
\mathcal{L}_{total} = \mathcal{L}_{det} + \mathcal{L}_{quality}.
  \label{Eq4}
\end{equation}

\begin{algorithm}[t!]
\SetAlgoLined
\DontPrintSemicolon
\SetNoFillComment
\footnotesize
\KwIn{A video sequence $\texttt{V}$; object detector $\texttt{Det}$; detection score threshold {$\tau$}; quality score threshold {$\mu_v$, $\mu_t$}}
\KwOut{Tracks $\mathcal{T}$ of the video}

Initialization: $\mathcal{T} \leftarrow \emptyset$\;
\For{frame $f_k$ in $\texttt{V}$}{
	\tcc{boxes \& scores}
	$\mathcal{D}_k \leftarrow \texttt{Det}(f_k)$ \;
	$\mathcal{D}_{high} \leftarrow \emptyset$ \;
	$\mathcal{D}_{low} \leftarrow \emptyset$ \;
	\tcc{\textbf{first gate}}
	\For{$d$ in $\mathcal{D}_k$}{
	\If{$d.score > \tau$}{
	$\mathcal{D}_{high} \leftarrow  \mathcal{D}_{high} \cup \{d\}$ \;
	}
	\Else{
	$\mathcal{D}_{low} \leftarrow  \mathcal{D}_{low} \cup \{d\}$ \;
	}
	}
	\tcc{predict location}
	\For{$t$ in $\mathcal{T}$}{
	$t \leftarrow \texttt{CV}(t)$ \;
	}
	\tcc{association with high scores}
	Associate $\mathcal{T}$ and $\mathcal{D}_{high}$ using \texttt{L2 distance}\;
	$\mathcal{D}_{remain} \leftarrow \text{remaining object boxes from } \mathcal{D}_{high}$ \;
	$\mathcal{T}_{remain} \leftarrow \text{remaining tracks from } \mathcal{T}$ \;
    \tcc{association with low scores}
	Associate $\mathcal{T}_{remain}$ and $\mathcal{D}_{low}$ using \texttt{L2 distance}\;
	$\mathcal{T}_{sec},\mathcal{D}_{sec} \leftarrow \text{matched pairs from $\mathcal{T}_{remain}$ ,$\mathcal{D}_{low}$}$ \;
	$\mathcal{T}_{re-remain} \leftarrow \text{remaining tracks from } \mathcal{T}_{remain}$ \;
	\tcc{\textbf{second gate}}
	\For{$t$, $d$ in $\mathcal{T}_{sec}$, $\mathcal{D}_{sec}$}{
	\If{$t.v_{score} < \mu_v$ \text{or} $d.l_{score} < \mu_t$}{
	$\mathcal{T}_{re-remain} \leftarrow  \mathcal{T}_{re-remain} \cup \{t\}$ \;
	}
	}
	\tcc{update and initialize}
	$\mathcal{T} \leftarrow \mathcal{T} \setminus \mathcal{T}_{re-remain}$ \;
    \For{$d$ in $\mathcal{D}_{remain}$}{
	$\mathcal{T} \leftarrow  \mathcal{T} \cup \{d\}$ \;
	}
}
Return: $\mathcal{T}$
\caption{Pseudo-code of QOA.}
\label{algo:qoa}
\end{algorithm}

\label{SOTAs}
\begin{table*}[t]
\label{tab:sota}
\centering
\scalebox{0.93}{
\begin{tabular}{l| c | c c c c c}
% \specialrule{1pt}{0pt}{1pt}
\toprule
Methods & Modality & AMOTA $\uparrow$ & AMOTP $\downarrow$ & RECALL $\uparrow$ & MOTA $\uparrow$ & IDS $\downarrow$ \\
\midrule
\multicolumn{7}{c}{Validation Split} \\
\midrule
CenterPoint~\cite{centerpoint}  & LiDAR & 0.665  & 0.567 & 69.9\% & 0.562 & 562\\
SimpleTrack~\cite{simpletrack}  & LiDAR & 0.687  & 0.573 & 72.5\% & 0.592 & 519\\
\midrule
DEFT~\cite{deft} & Camera & 0.201 & N/A & N/A &  0.171 & N/A\\
QD3DT~\cite{quasi-dense} & Camera & 0.242 & 1.518 & 39.9\% &  0.218 & 5646\\
TripletTrack~\cite{triplettrack} & Camera & 0.285 & 1.485 & N/A &  N/A & N/A\\
MUTR3D~\cite{mutr3d} & Camera & 0.294 & 1.498 & 42.7\%  & 0.267  & 3822\\
QTrack (Ours)  & Camera & \textbf{0.511} & \textbf{1.090} & \textbf{58.5\%}  & \textbf{0.465}  & \textbf{1144}\\
\midrule
\multicolumn{7}{c}{Test Split} \\
\midrule
CenterTrack~\cite{centertrack} & Camera & 0.046 & 1.543 & 23.3\% & 0.043 & 3807\\
% PermaTrack & Camera & 0.066 & 1.491 & 18.9\% & 0.06 & 3598 & \\
DEFT~\cite{deft} & Camera & 0.177 & 1.564 & 33.8\% & 0.156 &  6901\\
Time3D~\cite{time3d}  & Camera & 0.210 & 1.360 & N/A &  0.173 & N/A\\
QD3DT~\cite{quasi-dense} & Camera & 0.217 & 1.550 & 37.5\% &  0.198 & 6856 \\
TripletTrack~\cite{triplettrack} & Camera & 0.268 & 1.504 & 40.0\% &  0.245 & \textbf{1044}\\
MUTR3D~\cite{mutr3d} & Camera & 0.270 & 1.494 & 41.1\%  & 0.245  & 6018\\
PolarDETR~\cite{polardetr} & Camera & 0.273 & 1.185 & 40.4\%  & 0.238  & 2170\\
SRCN3D~\cite{srcn3d} & Camera & 0.398 & 1.317 & 53.8\%  & 0.359  & 4090\\
QTrack (Ours) & Camera & \textbf{0.480} & \textbf{1.107}& \textbf{56.9\%}  & \textbf{0.431}  & 1484\\
\bottomrule
%\specialrule{1pt}{1pt}{0pt}
\end{tabular}
}
\caption{Comparison with state-of-the-art methods on nuScenes validation and test dataset. Our QTrack employs VoVNet-99 initialized from DD3D as image backbone. The resolution of input image is 640 $\times$ 1600.}
%\label{tab:main} \vspace{-.5em}
\end{table*}

\begin{table*}[t]
\centering
\scalebox{0.85}{
\begin{tabular}{l|c |c c c c c c c c c}
% \specialrule{1pt}{0pt}{1pt}
\toprule
Methods & Backbone & AMOTA $\uparrow$ & AMOTP $\downarrow$ & RECALL $\uparrow$ & MOTA $\uparrow$ & MOTP $\downarrow$ & IDS $\downarrow$ & FRAG $\downarrow$ & MT $\uparrow$ & ML $\downarrow$\\
\midrule
BEVDepth + KF  & ResNet-50 & 0.303 & 1.337 & 39.7\% & 0.284 & 0.705 & 1290 & 780 & 1462 & 3344\\
BEVDepth + CV  & ResNet-50 & 0.325  & 1.276 & 42.8\% & 0.300 & 0.710 & 903 & 907 & 1843 & 3299\\
BEVDepth + SimpleTrack  & ResNet-50 & 0.338 & 1.294 & 43.9\% & 0.304 & 0.742 & 950 & 904 & 1798 & 3213\\
BEVDepth + Ours  & ResNet-50  & 0.347  & 1.347 & 42.6\% & 0.309 & 0.722 & 944 & 1106 & 1758 & 3137 \\
\midrule
BEVDepth + KF  & ResNet-101 & 0.301 & 1.345 & 40.2\% & 0.287 & 0.685 & 1444 & 841 & 1591 & 3156\\
BEVDepth + CV  & ResNet-101 & 0.323 & 1.282 & 42.1\% & 0.299 & 0.696 & 807 & 885 & 2359 & 3256 \\
BEVDepth + SimpleTrack  & ResNet-101 & 0.333 & 1.302 & 42.4\% & 0.303 & 0.701 & 887 & 904 & 1835 & 3174\\
BEVDepth + Ours  & ResNet-101 & 0.339  & 1.349 & 42.8\% & 0.309 & 0.691 & 1100 & 1187 & 1956 & 2890 \\
\bottomrule
%\specialrule{1pt}{1pt}{0pt}
\end{tabular}
}
\label{tab:track}
\caption{Comparison with different post-processing trackers on nuScenes validation dataset. We report the tracking results with two different backbones, and the resolution of the input image is 256 $\times$ 704.}
%\label{tab:main} \vspace{-.5em}
\end{table*}

The overall training procedure is an end-to-end manner while the quality prediction task will not damage the performance of the base detector. Moreover, the quality estimation is used in our proposed Quality-aware Object Association (QOA) module, which will be discussed next section.

\subsection{Quality-aware Object Association}

After obtaining the quality of the center location and velocity, we have more reference cues to achieve robust and accurate association. To this end, we propose a simple but effective quality-aware object association strategy (QOA). Specifically, QOA sets up two "gates". The first gate is the classification confidence score (cls score). We first separate the candidate detection boxes into high score ones and low score ones according to their cls scores. The high score candidates are first associated with the tracklets. Then the unmatched tracklets are associated with the low score candidates. These low score candidates are most caused by occlusion, motion blur, or light weakness, which are easily confused with the miscellaneous boxes. To deal with the issue, the second gate, quality uncertainty score, is introduced. After getting the second association results between the unmatched tracklets and the low score candidates, we then recheck the matched track-det pairs according to the location and velocity quality scores. Only high-quality matched track-det pairs can remain and low-quality pairs are regarded as the mismatch. The pseudo-code of QOA is shown in Algorithm 1.

Benefiting from the quality estimation, QOA does not need a complex motion or appearance model to provide association cues. A simple velocity prediction (CV) is enough (line \#15). Hence, we use the velocity of the tracklet at frame $t-1$ to predict the center location at frame $t$ and then compute the L2 distance between predictions and candidate detections (line \#17 and line \#20) as the similarity. At last, we apply the similarity with the Hungarian algorithm to get the association results. Mathematically, 

\begin{equation}
\begin{aligned}
&c_t = c_{t-1} + v_{t-1} \Delta t \\
&cost = \mathcal{L}_2(c_t, d_t) \\
&match = Hungarian(cost),
  \label{Eq5}
\end{aligned}
\end{equation}
where $c_{t-1}$, $v_{t-1}$ represents the center location and velocity of the tracklets at frame $t-1$. $d_t$ is the candidate detection center location at frame $t$ and $\Delta t$ is the time lag.

\section{Experiments}

\subsection{Datasets and Metrics}

\noindent \textbf{Datasets.} We mainly evaluate our QTrack on the 3D detection and tracking datasets of nuScenes. nuScenes dataset is a large-scale autonomous driving benchmark that consists of 1000 real-world sequences, 700 sequences for training, 150 for validation, and 150 for the test. Each sequence has roughly 40 keyframes, which are annotated by each sensor (e.g., LiDAR, Radar, and Camera) with a sampling rate of 2 FPS. Each frame includes images from six cameras with a full 360-degree field of view. For the detection task, there are 1.4 M annotated 3D bounding boxes from 10 categories. For the tracking task, it provides 3D tracking bounding boxes from 7 categories.

\noindent \textbf{Metrics.} For 3D detection task, we report nuScenes Detection Score (NDS), mean Average Prediction (mAP), as well as five True Positive (TP) metrics including mean Average Translation Error (mATE), mean Average Scale Error (mASE), mean Average Orientation Error (mAOE), mean Average Velocity Error (mAVE), mean Average Attribute Error (mAAE). For 3D tracking task, we report Average Multi-object Tracking Accuracy (AMOTA) and Average Multi-object Tracking Precision (AMOTP). We also report metrics used in 2D tracking task from CLEAR \cite{bernardin2006multiple}, e.g., MOTA, MOTP, and IDS.

\subsection{Implementation Details}

Following BEVdepth, we adopt three types of backbone: ResNet-50~\cite{resnet}, ResNet-101, and VoVNet-99 (Initialized from DD3D~\cite{dd3d}) as the image backbone. If not specified, the image size is processed to $256 \times 704$. The data augmentation includes random cropping, random scaling, random flipping, and random rotation. In addition, we also adopt BEV data augmentations including random scaling, random flipping, and random rotation. We use AdamW as optimizer with learning rate of $2 \times 10 ^{-4}$ and batch size of 64. When compared with other methods, QTrack is trained for 24 epochs for ResNet and 20 epochs for VoVNet with CBGS~\cite{cbgs}.

\subsection{Comparision with Preceding SOTAs}

\noindent \textbf{Test and validation set.} We compare the performance of QTrack with preceding SOTA methods on the nuScenes benchmark. The results are reported in Tab.~1. Our QTrack outperforms all current SOTA methods for the camera-based trackers by a large margin. For both validation and test sets, all reported metrics (e.g., AMOTA, AMOTP, RECALL, IDS, etc.) achieve best performance. Specially, AMOTA result of QTrack first achieves 0.511, which significantly reduces the performance gap between the pure camera and LiDAR-based trackers.

\noindent \textbf{Compare with other post-processing trackers.} Tab.~2 illustrates that QTrack outperforms the naive Kalman filter based method and its advanced variant from SimpleTrack~\cite{simpletrack} by employing identical 3D detector and backbone settings. Moreover, our method only needs simple operations (i.e., Matrix multiplication and addition) for tracking procedure, while Kalman filter based ones need relatively complex operation like matrix transpose and the complex process for adjusting hyper-parameters. 
The overall tracking framework is significantly efficient and will not trigger a serious latency, which is fatal in a real perception scenario~\cite{streamyolo1,streamyolo2}.

\begin{table}[t!]
\centering
\scalebox{0.73}{
\begin{tabular}{c|c|cc|cccc}
    \toprule
	  Backbone& MF & CLS & Q. & AMOTA$\uparrow$ & AMOTP$\downarrow$ & MOTA$\uparrow$ & IDS$\downarrow$ \\ 
    \midrule
	  \multirow{6}{*}{ResNet-50}& \multirow{3}{*}{\XSolidBrush} &  &  & 29.1 & 1.314 & 26.7 & 1488 \\
	  &  & \Checkmark &  & 30.7 & 1.394 & 28.3 & 1748 \\
	  &  & \Checkmark & \Checkmark & 31.3 & 1.390 & 28.5 & 1559\\
	 \cmidrule{2-8}
	  & \multirow{3}{*}{\Checkmark} &  & & 32.5 & 1.276 & 30.0 & 903\\
	  &  & \Checkmark && 34.1 & 1.348 & 30.5 & 1141 \\
	  &  & \Checkmark & \Checkmark & 34.7 & 1.347 & 30.9 & 944\\
	 \midrule
	  \multirow{6}{*}{ResNet-101}& \multirow{3}{*}{\XSolidBrush} &  &  & 29.1 & 1.314 & 26.7 & 1488 \\
	  &  & \Checkmark &  & 31.2 & 1.389 & 28.4 & 1622 \\
	  &  & \Checkmark & \Checkmark & 31.8 & 1.386 & 29.1 & 1638\\
	 \cmidrule{2-8}
	  & \multirow{3}{*}{\Checkmark} &  & & 32.3 & 1.282 & 30.9 & 1100\\
	  &  & \Checkmark & & 33.2 & 1.352 & 30.3 & 1053 \\
	  &  & \Checkmark & \Checkmark & 33.9 & 1.349 & 30.9 & 1100\\
	 \midrule
	 \multirow{6}{*}{VoVNet-99} & \multirow{3}{*}{\XSolidBrush} &  &  & 38.8 & 1.220 & 35.3 & 1670\\
	 &  & \Checkmark & & 40.4 & 1.266 & 36.9 & 1575\\
	 &  & \Checkmark & \Checkmark & 40.8 & 1.259 & 37.0 & 1527 \\
	 \cmidrule{2-8}
	 & \multirow{3}{*}{\Checkmark} & & & 41.7 & 1.177 & 37.3 & 914\\
	 &  & \Checkmark && 42.2 & 1.236 & 38.1 & 1125 \\
	 & & \Checkmark & \Checkmark &  42.6 & 1.228 & 38.3 & 1076\\
    \bottomrule
\end{tabular}}
\caption{Ablation study of the components in QTrack. CLS indicates the first gate classification score while Q. indicates the second gate, i.e., quality score.}
\label{tab:components}
% \vspace{-1mm}
\end{table}

\begin{table}[t!]
\centering
\scalebox{0.9}{
\begin{tabular}{c|cc|cccc}
    \toprule
	  MF & VQ & LQ & AMOTA$\uparrow$ & AMOTP$\downarrow$ & MOTA$\uparrow$ & IDS$\downarrow$ \\ 
    \midrule
	  \multirow{4}{*}{\XSolidBrush} &  &  & 40.4 & 1.266 & 36.9 & 1575 \\
	   & \Checkmark &  & 40.1 & 1.269 & 36.6 & 1680 \\
	   &  & \Checkmark & 40.5 & 1.264 & \bf{37.3} & \bf{1445} \\
	  & \Checkmark & \Checkmark & \bf{40.8} & \bf{1.259} & 37.0 & 1527\\
	 \midrule
	  \multirow{4}{*}{\Checkmark} &  & & 42.2 & 1.236 & 38.1 & 1125\\
	   & \Checkmark && 41.9 & 1.239 & 38.0 & \bf{999} \\
	   &  & \Checkmark & 42.2 & 1.235 & 38.2 & 1023 \\
	   & \Checkmark & \Checkmark & \bf{42.6} & \bf{1.228} & \bf{38.3} & 1076\\
    \bottomrule
\end{tabular}}	 
  \caption{Ablation study of how to use velocity quality (VQ) and location quality (LQ).}
  \label{tab:quality}
\end{table}

\subsection{Ablation Study}
\label{ablation}
In this subsection, we verify the effectiveness of the proposed strategies separately through ablation studies. All the experiments are conducted on the nuScenes \emph{val} set. 
%All models are trained for 24 epochs.

\noindent \textbf{Analysis of the components of QTrack.}  In this part, we verify the effectiveness of various components in QTrack through an ablation study. As shown in Tab.~\ref{tab:components}, the first row of the table shows baseline performance for tracking when using BEVDepth detections followed by a simple velocity association step (CV method). We can observe that the two gates of QOA can both develop the tracking performance in the all settings (ResNet-50, ResNet-101 or VoVNet-99, single-frame or multi-frame), which means that the filter for the low-quality association results is necessary. Furthermore, we can observe that the metric of IDS increases when applying the first gate by classification confidence score. This phenomenon shows that only considering confidence score inevitably introduces low-quality bounding boxes, which causes bad association cases. Therefore, the second gate, quality score, can provide a fine-grained reference to achieve a better association trade-off.

\noindent \textbf{Analysis of the location and velocity quality for tracking.} In this part, we conduct an in-depth analysis on the location and velocity quality score for the association process. As mentioned before, location and velocity quality scores are obtained by the quality branch. Then they are both regarded as the reference clues to filter the low classification confidence association results in QOA. We verify the performance of only using one of them as the second gate of QOA, and the results are reported in Tab.~\ref{tab:quality}. As shown, only using one of the location and velocity quality scores does not contribute to the tracking performance, which confirms our analysis that the location and velocity quality is not aligned and we should take both of them into consideration.

\begin{table}[t]
\centering
\scalebox{0.91}{
\begin{tabular}{l|cc|cc}
\toprule
Extra Branch & mAP$\uparrow$ & NDS$\uparrow$  &  CV & SimpleTrack\\
\midrule
None & 0.3579 & 0.4826 & 0.326 & 0.337\\
\midrule
Appearance & 0.3522 & 0.4798 & 0.315 & 0.328\\
Relative Drop & -0.57\% & -0.38\% & -1.1\% & -0.9\%\\
\midrule
Quality & 0.3585 & 0.4831 & 0.325 & 0.338\\
Relative Drop & \textbf{+0.06\%} & \textbf{+0.05\%} & -0.1\% & \textbf{+0.1\%}\\
\bottomrule
\end{tabular}} 
\caption{Influence of extra branch on performance and tracking detection. For tracking performance of CV and SimpleTrack, we report the AMOTA metric for comparison.}
\label{tab:detector}
% \vspace{-1mm}
\end{table} 

\noindent \textbf{Influence on base 3D detector.} As shown in Tab.~5, it proves that adding quality prediction branch does not affect the performance of base 3D detector. This is an extremely important property since post-processing trackers normally rely on the super performance of detector. Going one step further, we report the tracking performance by employing existing CV and SimpleTrack scheme. It reveals that tracking performance will not be affected by our quality branch, which agree with our designing purpose of Sec.~1. Then, we explore to append a appearance branch for extracting instance wised appearance embedding, which implement is the same as \cite{fairmot}. The results show that slight performance degradation (nearly 0.5\%) is triggered on detection task, but it significantly damages the performance of tracking task by nearly 1.0\%. It reflects that our method is more effective and efficient.

\subsection{Discussion and Future Work}

Inspired by \cite{iounet,iouaware,scd}, we explore to incorporate velocity quality $V$ with classification score $C$ as $M$, which is adopted to act as threshold metric in NMS procedure. Technically, we formulate $M$ in Eq. 6, in which $\alpha$ is a hyper-parameter to control the contribution of $V$.

\begin{equation}
M = V^{(1-\alpha)} \cdot C^\alpha,
  \label{Eq6}
\end{equation}

As shown in Fig. 4, we plot the four performance metrics of detection task by controlling $\alpha$. It reflects that as contribution of $V$ becomes bigger, mAVE drops dramatically. However, it also brings about inevitable performance degradation for mAP and mATE metrics. NDS, as a comprehensive metric, becomes better and then gets worse as $\alpha$ changes larger, which is actually a trade-off between location error and velocity error. This phenomenon agrees with our viewpoint in Sec.~1, i.e., the quality of these two prediction tasks are not align. Combining the performance of detection and tracking tasks with respect to above imbalance issue, it exposes a challenge: \emph{how to design a method to simultaneously predict location (or 3D box) and velocity well?} This challenge can help further boost performance of 3D detection task or other downstream tasks like 3D MOT.

\begin{figure}[t]
\centering
\includegraphics[width=1.0\columnwidth]{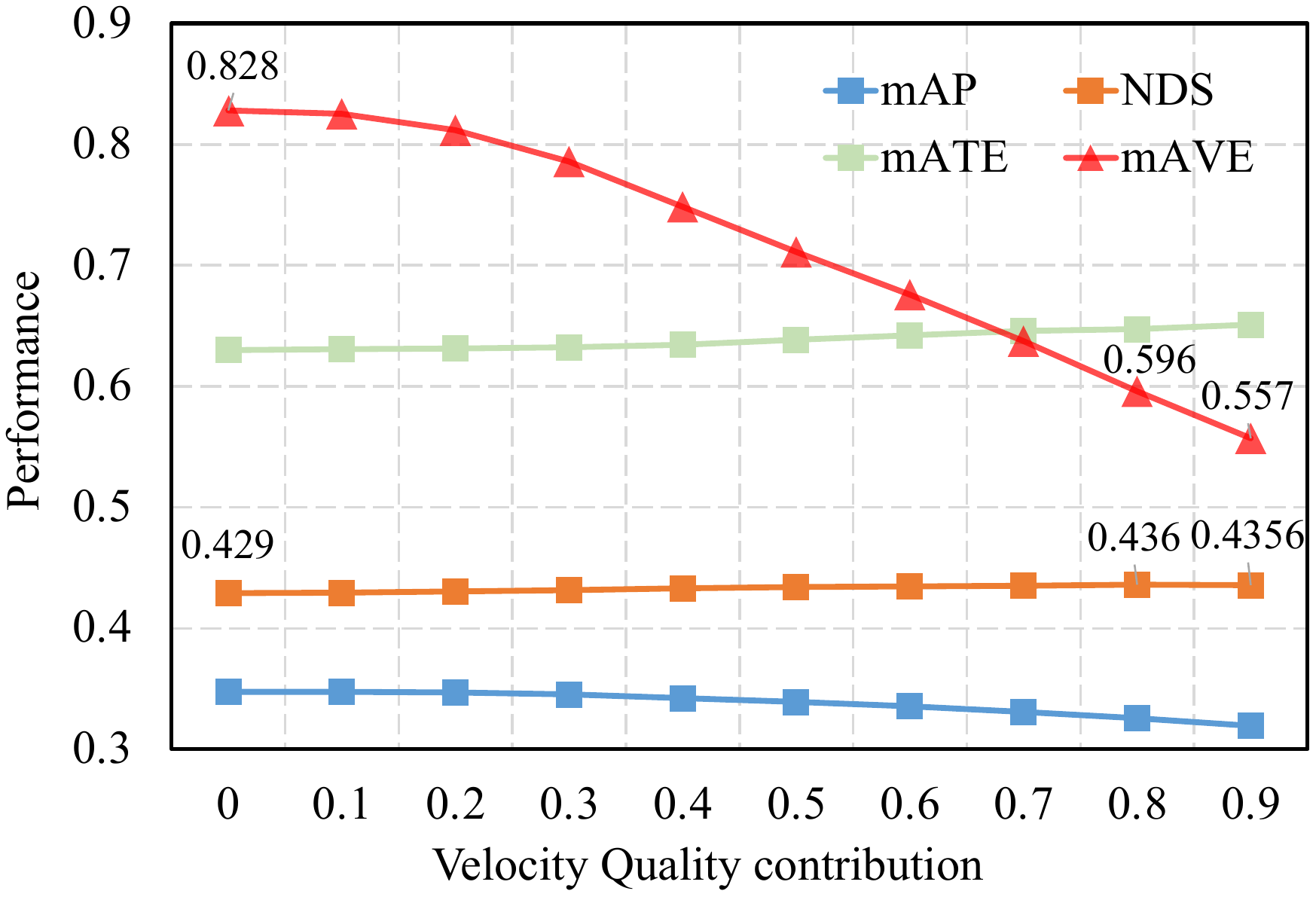}
\caption{Influence of velocity quality on detection performance in NMS procedure. The abscissa indicates $\alpha$ in Eq. 6.}
\label{fig:nms}
\end{figure}

% \subsection{Limitation}

% We revealed the impact of imbalance issue between location and velocity quality for 3D MOT and 3D detection tasks. To this end, we only used a simple supervision to learning their qualities. We believe there are better schemes to implementation, where uncertainty based manners~\cite{gaussianyolo,klloss} may be worked. In addition, how to design a better quality metric is a promising point. Moreover, it may be also a breakthrough to tackle the imbalance issue by improve base detector.

\section{Conclusion}
\label{conclusion}

In this paper, we analyze the imbalance prediction quality distribution of location and velocity. It motivates us to propose a Quality-aware Object Association (QOA) method to alleviate the imbalance issue for 3D multi-object tracking (3D MOT). To this end, we introduce Normalized Gaussian Quality (NGQ) metric to measure the predicted quality of location and velocity, and structure an effective module for quality learning. Afterwards, we further present QTrack, an “tracking by detection” framework for 3D MOT in multi-view camera scene, which incorporats with QOA to perform tracking procedure. The extensive experiments demonstrate the efficacy and robustness of our method. Finally, we release a challenge to inspire more research to focus on the imbalance between localization and velocity qualities for both 3D detection and tracking tasks.

\clearpage

{\small
\bibliography{aaai23}

\begin{thebibliography}{43}
\providecommand{\natexlab}[1]{#1}

\bibitem[{Bernardin, Elbs, and Stiefelhagen(2006)}]{bernardin2006multiple}
Bernardin, K.; Elbs, A.; and Stiefelhagen, R. 2006.
\newblock Multiple object tracking performance metrics and evaluation in a
  smart room environment.
\newblock In \emph{Sixth IEEE International Workshop on Visual Surveillance, in
  conjunction with ECCV}, volume~90. Citeseer.

\bibitem[{Caesar et~al.(2020)Caesar, Bankiti, Lang, Vora, Liong, Xu, Krishnan,
  Pan, Baldan, and Beijbom}]{nuscenes}
Caesar, H.; Bankiti, V.; Lang, A.~H.; Vora, S.; Liong, V.~E.; Xu, Q.; Krishnan,
  A.; Pan, Y.; Baldan, G.; and Beijbom, O. 2020.
\newblock nuscenes: A multimodal dataset for autonomous driving.
\newblock In \emph{Proceedings of the IEEE/CVF conference on computer vision
  and pattern recognition}, 11621--11631.

\bibitem[{Chaabane et~al.(2021)Chaabane, Zhang, Beveridge, and O'Hara}]{deft}
Chaabane, M.; Zhang, P.; Beveridge, J.~R.; and O'Hara, S. 2021.
\newblock Deft: Detection embeddings for tracking.
\newblock \emph{arXiv preprint arXiv:2102.02267}.

\bibitem[{Chen et~al.(2022)Chen, Wang, Cheng, Zhang, Huang, and
  Liu}]{polardetr}
Chen, S.; Wang, X.; Cheng, T.; Zhang, Q.; Huang, C.; and Liu, W. 2022.
\newblock Polar parametrization for vision-based surround-view 3d detection.
\newblock \emph{arXiv preprint arXiv:2206.10965}.

\bibitem[{Han et~al.(2022)Han, Huang, Wang, Yu, Liu, and Pan}]{han2022mat}
Han, S.; Huang, P.; Wang, H.; Yu, E.; Liu, D.; and Pan, X. 2022.
\newblock Mat: Motion-aware multi-object tracking.
\newblock \emph{Neurocomputing}, 476: 75--86.

\bibitem[{He et~al.(2016)He, Zhang, Ren, and Sun}]{resnet}
He, K.; Zhang, X.; Ren, S.; and Sun, J. 2016.
\newblock Deep residual learning for image recognition.
\newblock In \emph{Proceedings of the IEEE conference on computer vision and
  pattern recognition}, 770--778.

\bibitem[{Hu et~al.(2022)Hu, Yang, Fischer, Darrell, Yu, and Sun}]{quasi-dense}
Hu, H.-N.; Yang, Y.-H.; Fischer, T.; Darrell, T.; Yu, F.; and Sun, M. 2022.
\newblock Monocular quasi-dense 3d object tracking.
\newblock \emph{IEEE Transactions on Pattern Analysis and Machine
  Intelligence}.

\bibitem[{Huang et~al.(2021)Huang, Huang, Zhu, and Du}]{huang2021bevdet}
Huang, J.; Huang, G.; Zhu, Z.; and Du, D. 2021.
\newblock Bevdet: High-performance multi-camera 3d object detection in
  bird-eye-view.
\newblock \emph{arXiv preprint arXiv:2112.11790}.

\bibitem[{Huang et~al.(2019)Huang, Huang, Gong, Huang, and Wang}]{maskscoring}
Huang, Z.; Huang, L.; Gong, Y.; Huang, C.; and Wang, X. 2019.
\newblock Mask scoring r-cnn.
\newblock In \emph{Proceedings of the IEEE/CVF conference on computer vision
  and pattern recognition}, 6409--6418.

\bibitem[{Jiang et~al.(2018)Jiang, Luo, Mao, Xiao, and Jiang}]{iounet}
Jiang, B.; Luo, R.; Mao, J.; Xiao, T.; and Jiang, Y. 2018.
\newblock Acquisition of localization confidence for accurate object detection.
\newblock In \emph{Proceedings of the European conference on computer vision
  (ECCV)}, 784--799.

\bibitem[{Kalman(1960)}]{kalman}
Kalman, R.~E. 1960.
\newblock A new approach to linear filtering and prediction problems.
\newblock \emph{Transactions of the ASME–Journal of Basic Engineering
  82(Series D)}, 35--45.

\bibitem[{Lang et~al.(2019)Lang, Vora, Caesar, Zhou, Yang, and
  Beijbom}]{pointpillars}
Lang, A.~H.; Vora, S.; Caesar, H.; Zhou, L.; Yang, J.; and Beijbom, O. 2019.
\newblock Pointpillars: Fast encoders for object detection from point clouds.
\newblock In \emph{Proceedings of the IEEE/CVF conference on computer vision
  and pattern recognition}, 12697--12705.

\bibitem[{Li and Jin(2022)}]{time3d}
Li, P.; and Jin, J. 2022.
\newblock Time3D: End-to-End Joint Monocular 3D Object Detection and Tracking
  for Autonomous Driving.
\newblock In \emph{Proceedings of the IEEE/CVF Conference on Computer Vision
  and Pattern Recognition}, 3885--3894.

\bibitem[{Li et~al.(2022{\natexlab{a}})Li, Ge, Yu, Yang, Wang, Shi, Sun, and
  Li}]{li2022bevdepth}
Li, Y.; Ge, Z.; Yu, G.; Yang, J.; Wang, Z.; Shi, Y.; Sun, J.; and Li, Z.
  2022{\natexlab{a}}.
\newblock BEVDepth: Acquisition of Reliable Depth for Multi-view 3D Object
  Detection.
\newblock \emph{arXiv preprint arXiv:2206.10092}.

\bibitem[{Li et~al.(2022{\natexlab{b}})Li, Qu, Zhou, Liu, Wang, and
  Jiang}]{li2022diversity}
Li, Z.; Qu, Z.; Zhou, Y.; Liu, J.; Wang, H.; and Jiang, L. 2022{\natexlab{b}}.
\newblock Diversity Matters: Fully Exploiting Depth Clues for Reliable
  Monocular 3D Object Detection.
\newblock In \emph{Proceedings of the IEEE/CVF Conference on Computer Vision
  and Pattern Recognition}, 2791--2800.

\bibitem[{Li et~al.(2022{\natexlab{c}})Li, Wang, Li, Xie, Sima, Lu, Yu, and
  Dai}]{bevformer}
Li, Z.; Wang, W.; Li, H.; Xie, E.; Sima, C.; Lu, T.; Yu, Q.; and Dai, J.
  2022{\natexlab{c}}.
\newblock BEVFormer: Learning Bird's-Eye-View Representation from Multi-Camera
  Images via Spatiotemporal Transformers.
\newblock \emph{arXiv preprint arXiv:2203.17270}.

\bibitem[{Liu et~al.(2022)Liu, Wang, Zhang, and Sun}]{liu2022petr}
Liu, Y.; Wang, T.; Zhang, X.; and Sun, J. 2022.
\newblock Petr: Position embedding transformation for multi-view 3d object
  detection.
\newblock \emph{arXiv preprint arXiv:2203.05625}.

\bibitem[{Marinello, Proesmans, and Van~Gool(2022)}]{triplettrack}
Marinello, N.; Proesmans, M.; and Van~Gool, L. 2022.
\newblock TripletTrack: 3D Object Tracking Using Triplet Embeddings and LSTM.
\newblock In \emph{Proceedings of the IEEE/CVF Conference on Computer Vision
  and Pattern Recognition}, 4500--4510.

\bibitem[{Pang, Li, and Wang(2021)}]{simpletrack}
Pang, Z.; Li, Z.; and Wang, N. 2021.
\newblock Simpletrack: Understanding and rethinking 3d multi-object tracking.
\newblock \emph{arXiv preprint arXiv:2111.09621}.

\bibitem[{Park et~al.(2021)Park, Ambrus, Guizilini, Li, and Gaidon}]{dd3d}
Park, D.; Ambrus, R.; Guizilini, V.; Li, J.; and Gaidon, A. 2021.
\newblock Is pseudo-lidar needed for monocular 3d object detection?
\newblock In \emph{Proceedings of the IEEE/CVF International Conference on
  Computer Vision}, 3142--3152.

\bibitem[{Shi et~al.(2020)Shi, Guo, Jiang, Wang, Shi, Wang, and Li}]{pvrcnn}
Shi, S.; Guo, C.; Jiang, L.; Wang, Z.; Shi, J.; Wang, X.; and Li, H. 2020.
\newblock Pv-rcnn: Point-voxel feature set abstraction for 3d object detection.
\newblock In \emph{Proceedings of the IEEE/CVF Conference on Computer Vision
  and Pattern Recognition}, 10529--10538.

\bibitem[{Shi, Wang, and Li(2019)}]{pointrcnn}
Shi, S.; Wang, X.; and Li, H. 2019.
\newblock Pointrcnn: 3d object proposal generation and detection from point
  cloud.
\newblock In \emph{Proceedings of the IEEE/CVF conference on computer vision
  and pattern recognition}, 770--779.

\bibitem[{Shi et~al.(2022)Shi, Shen, Sun, Wang, Li, Sun, Jiang, and
  Yang}]{srcn3d}
Shi, Y.; Shen, J.; Sun, Y.; Wang, Y.; Li, J.; Sun, S.; Jiang, K.; and Yang, D.
  2022.
\newblock SRCN3D: Sparse R-CNN 3D Surround-View Camera Object Detection and
  Tracking for Autonomous Driving.
\newblock \emph{arXiv preprint arXiv:2206.14451}.

\bibitem[{Tian et~al.(2019)Tian, Shen, Chen, and He}]{fcos}
Tian, Z.; Shen, C.; Chen, H.; and He, T. 2019.
\newblock Fcos: Fully convolutional one-stage object detection.
\newblock In \emph{Proceedings of the IEEE/CVF international conference on
  computer vision}, 9627--9636.

\bibitem[{Vaswani et~al.(2017)Vaswani, Shazeer, Parmar, Uszkoreit, Jones,
  Gomez, Kaiser, and Polosukhin}]{transformer}
Vaswani, A.; Shazeer, N.; Parmar, N.; Uszkoreit, J.; Jones, L.; Gomez, A.~N.;
  Kaiser, {\L}.; and Polosukhin, I. 2017.
\newblock Attention is all you need.
\newblock \emph{Advances in neural information processing systems}, 30.

\bibitem[{Wang et~al.(2021)Wang, Zhu, Pang, and Lin}]{fcos3d}
Wang, T.; Zhu, X.; Pang, J.; and Lin, D. 2021.
\newblock Fcos3d: Fully convolutional one-stage monocular 3d object detection.
\newblock In \emph{Proceedings of the IEEE/CVF International Conference on
  Computer Vision}, 913--922.

\bibitem[{Wang et~al.(2022)Wang, Guizilini, Zhang, Wang, Zhao, and
  Solomon}]{detr3d}
Wang, Y.; Guizilini, V.~C.; Zhang, T.; Wang, Y.; Zhao, H.; and Solomon, J.
  2022.
\newblock Detr3d: 3d object detection from multi-view images via 3d-to-2d
  queries.
\newblock In \emph{Conference on Robot Learning}, 180--191. PMLR.

\bibitem[{Weng et~al.(2020{\natexlab{a}})Weng, Wang, Held, and Kitani}]{ab3d}
Weng, X.; Wang, J.; Held, D.; and Kitani, K. 2020{\natexlab{a}}.
\newblock 3d multi-object tracking: A baseline and new evaluation metrics.
\newblock In \emph{2020 IEEE/RSJ International Conference on Intelligent Robots
  and Systems (IROS)}, 10359--10366. IEEE.

\bibitem[{Weng et~al.(2020{\natexlab{b}})Weng, Wang, Man, and
  Kitani}]{gnn3dmot}
Weng, X.; Wang, Y.; Man, Y.; and Kitani, K.~M. 2020{\natexlab{b}}.
\newblock Gnn3dmot: Graph neural network for 3d multi-object tracking with
  2d-3d multi-feature learning.
\newblock In \emph{Proceedings of the IEEE/CVF Conference on Computer Vision
  and Pattern Recognition}, 6499--6508.

\bibitem[{Wu, Li, and Wang(2020)}]{iouaware}
Wu, S.; Li, X.; and Wang, X. 2020.
\newblock IoU-aware single-stage object detector for accurate localization.
\newblock \emph{Image and Vision Computing}, 97: 103911.

\bibitem[{Yan, Mao, and Li(2018)}]{second}
Yan, Y.; Mao, Y.; and Li, B. 2018.
\newblock Second: Sparsely embedded convolutional detection.
\newblock \emph{Sensors}, 18(10): 3337.

\bibitem[{Yang et~al.(2022{\natexlab{a}})Yang, Liu, Li, Li, and
  Sun}]{streamyolo1}
Yang, J.; Liu, S.; Li, Z.; Li, X.; and Sun, J. 2022{\natexlab{a}}.
\newblock Real-time Object Detection for Streaming Perception.
\newblock In \emph{Proceedings of the IEEE/CVF Conference on Computer Vision
  and Pattern Recognition}, 5385--5395.

\bibitem[{Yang et~al.(2022{\natexlab{b}})Yang, Liu, Li, Li, and
  Sun}]{streamyolo2}
Yang, J.; Liu, S.; Li, Z.; Li, X.; and Sun, J. 2022{\natexlab{b}}.
\newblock StreamYOLO: Real-time Object Detection for Streaming Perception.
\newblock \emph{arXiv preprint arXiv:2207.10433}.

\bibitem[{Yang et~al.(2022{\natexlab{c}})Yang, Song, Liu, Li, Li, Sun, Sun, and
  Zheng}]{dbq}
Yang, J.; Song, L.; Liu, S.; Li, Z.; Li, X.; Sun, H.; Sun, J.; and Zheng, N.
  2022{\natexlab{c}}.
\newblock DBQ-SSD: Dynamic Ball Query for Efficient 3D Object Detection.
\newblock \emph{arXiv preprint arXiv:2207.10909}.

\bibitem[{Yang et~al.(2022{\natexlab{d}})Yang, Wu, Gou, Yu, Lin, Wang, Wang,
  Li, and Li}]{scd}
Yang, J.; Wu, S.; Gou, L.; Yu, H.; Lin, C.; Wang, J.; Wang, P.; Li, M.; and Li,
  X. 2022{\natexlab{d}}.
\newblock SCD: A stacked carton dataset for detection and segmentation.
\newblock \emph{Sensors}, 22(10): 3617.

\bibitem[{Yin, Zhou, and Krahenbuhl(2021)}]{centerpoint}
Yin, T.; Zhou, X.; and Krahenbuhl, P. 2021.
\newblock Center-based 3d object detection and tracking.
\newblock In \emph{Proceedings of the IEEE/CVF conference on computer vision
  and pattern recognition}, 11784--11793.

\bibitem[{Yu, Li, and Han(2022)}]{yu2022towards}
Yu, E.; Li, Z.; and Han, S. 2022.
\newblock Towards Discriminative Representation: Multi-view Trajectory
  Contrastive Learning for Online Multi-object Tracking.
\newblock In \emph{Proceedings of the IEEE/CVF Conference on Computer Vision
  and Pattern Recognition}, 8834--8843.

\bibitem[{Yu et~al.(2022)Yu, Li, Han, and Wang}]{yu2022relationtrack}
Yu, E.; Li, Z.; Han, S.; and Wang, H. 2022.
\newblock Relationtrack: Relation-aware multiple object tracking with decoupled
  representation.
\newblock \emph{IEEE Transactions on Multimedia}.

\bibitem[{Zhang et~al.(2022{\natexlab{a}})Zhang, Chen, Wang, Wang, and
  Zhao}]{mutr3d}
Zhang, T.; Chen, X.; Wang, Y.; Wang, Y.; and Zhao, H. 2022{\natexlab{a}}.
\newblock MUTR3D: A Multi-camera Tracking Framework via 3D-to-2D Queries.
\newblock In \emph{Proceedings of the IEEE/CVF Conference on Computer Vision
  and Pattern Recognition}, 4537--4546.

\bibitem[{Zhang et~al.(2022{\natexlab{b}})Zhang, Sun, Jiang, Yu, Weng, Yuan,
  Luo, Liu, and Wang}]{bytetrack}
Zhang, Y.; Sun, P.; Jiang, Y.; Yu, D.; Weng, F.; Yuan, Z.; Luo, P.; Liu, W.;
  and Wang, X. 2022{\natexlab{b}}.
\newblock ByteTrack: Multi-Object Tracking by Associating Every Detection Box.
\newblock In \emph{Proceedings of the European Conference on Computer Vision
  (ECCV)}.

\bibitem[{Zhang et~al.(2021)Zhang, Wang, Wang, Zeng, and Liu}]{fairmot}
Zhang, Y.; Wang, C.; Wang, X.; Zeng, W.; and Liu, W. 2021.
\newblock Fairmot: On the fairness of detection and re-identification in
  multiple object tracking.
\newblock \emph{International Journal of Computer Vision}, 129(11): 3069--3087.

\bibitem[{Zhou, Koltun, and Kr{\"a}henb{\"u}hl(2020)}]{centertrack}
Zhou, X.; Koltun, V.; and Kr{\"a}henb{\"u}hl, P. 2020.
\newblock Tracking objects as points.
\newblock In \emph{European Conference on Computer Vision}, 474--490. Springer.

\bibitem[{Zhu et~al.(2019)Zhu, Jiang, Zhou, Li, and Yu}]{cbgs}
Zhu, B.; Jiang, Z.; Zhou, X.; Li, Z.; and Yu, G. 2019.
\newblock Class-balanced grouping and sampling for point cloud 3d object
  detection.
\newblock \emph{arXiv preprint arXiv:1908.09492}.

\end{thebibliography}
}
\end{document}